# Image Translation-Based Unsupervised Cross-Modality Domain Adaptation for Medical Image Segmentation

Tao Yang[1][0000-0003-1440-0390] and Lisheng Wang[1][0000−0003−3234−7511]

[1] Department of Automation, Shanghai Jiao Tong University, Shanghai, China
{yangtao22,lswang}@sjtu.edu.cn

**Abstract.** Supervised deep learning usually faces more challenges in medical images than in natural images. Since annotations in medical images require the expertise of doctors and are more time-consuming and expensive. Thus, some researchers turn to unsupervised learning methods, which usually face inevitable performance drops. In addition, medical images may have been acquired at different medical centers with different scanners and under different image acquisition protocols, so the modalities of the medical images are often inconsistent. This modality difference (domain shift) also reduces the applicability of deep learning methods. In this regard, we propose an unsupervised cross-modality domain adaptation method based on image translation by transforming the source modality image with annotation into the unannotated target modality and using its annotation to achieve supervised learning of the target modality. In addition, the subtle differences between translated pseudo images and real images are overcome by self-training methods to further improve the task performance of deep learning. The proposed method showed mean Dice Similarity Coefficient (DSC) and Average Symmetric Surface Distance (ASSD) of 0.8351 ± 0.1152 and 1.6712 ± 2.1948 for vestibular schwannoma (VS), 0.8098 ± 0.0233 and 0.2317 ± 0.1577 for cochlea on the VS and cochlea segmentation task of the Cross-Modality Domain Adaptation (crossMoDA 2022) challenge validation phase leaderboard.

## 1 Introduction

Unsupervised domain adaptation (UDA) has attracted considerable interest due to the labor-intensive and expensive nature of medical data annotation, coupled with domain shifts across different imaging modalities. Despite its significance, the medical field lacks large-scale benchmarks to assess UDA methods. The crossMoDA benchmark is the first extensive multi-class dataset developed for unsupervised cross-modality domain adaptation [1-3]. In the crossMoDA 2022 challenge, the segmentation task focuses on delineating two critical brain structures (the tumor and cochlea) essential for vestibular schwannoma (VS) follow-up and treatment planning. Although contrast-enhanced T1 (ceT1) magnetic resonance imaging (MRI) scans are commonly used for VS segmentation, recent studies indicate that non-contrast sequences, such as high-resolution T2 (hrT2) imaging, can reduce risks associated with gadolinium-based agents and are more cost-effective. Accordingly, the challenge aims to perform VS and cochlea segmentation on unlabeled hrT2 scans using only labeled ceT1 scans within an unsupervised domain adaptation framework. To address this, we propose a cross-modality adaptation method that translates annotated source images into the target modality, leveraging their annotations for supervised learning. Furthermore, self-training techniques are employed to mitigate subtle discrepancies between the translated pseudo images and real images, thereby enhancing performance.

## 2    Methods

Our method consists of two parts: image translation and self-training, and the overall framework is shown in **Fig. 1**.

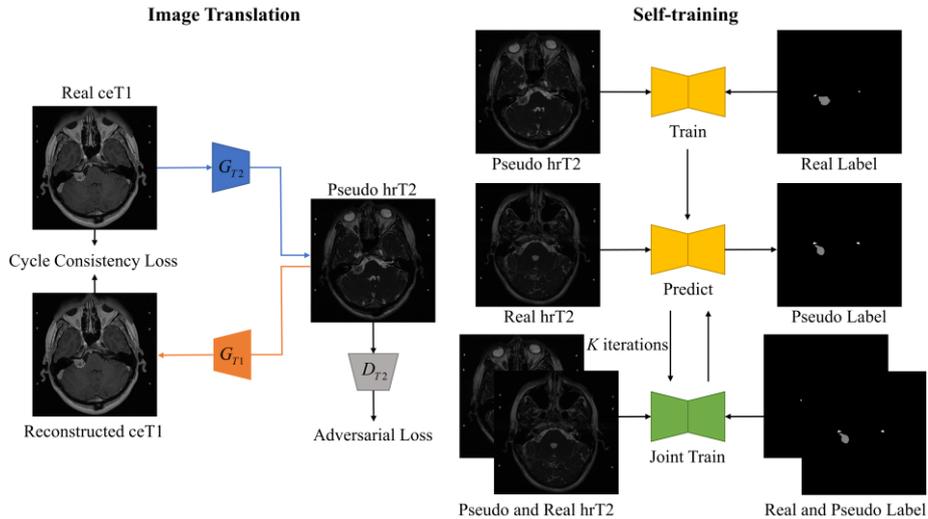

**Fig. 1.** The framework of unsupervised cross-modality domain adaptation for medical image segmentation based on image translation.

As illustrated in **Fig. 1**, an adversarial network is employed to translate source domain images (Real ceT1) into the target domain modality (Real hrT2), thereby generating pseudo target domain images (Pseudo hrT2). The self-training process unfolds as follows: (1) Supervised Training: Initially, a segmentation model is trained using the pseudo hrT2 images paired with real labels (Real Label). (2) Pseudo Label Generation: The trained model then segments the real hrT2 images, producing pseudo labels (Pseudo Label). (3) Joint Retraining: The segmentation network is retrained using a combined dataset comprising both the pseudo hrT2 images (with real labels) and the real hrT2 images (with pseudo labels). (4) This cycle of (2) generating pseudo labels and (3) joint retraining is repeated iteratively, continually refining the segmentation performance.

### 2.1 Image Translation

Since the source and target domain images provided by the crossMoDA 2022 challenge are unpaired, we employ Cycle-consistent adversarial networks (CycleGAN) [4] to translate annotated ceT1 scans into hrT2 scans. CycleGAN demonstrated its effectiveness in bridging the gap between these modalities during the crossMoDA 2021 challenge [1, 5, 6]. As CycleGAN is typically designed for 2D images, all 3D scans (both ceT1 and hrT2) are sliced along the z-axis to generate 2D images. Due to scanner differences, the original 3D volumes vary in size, resulting in inconsistent 2D slice dimensions. To preserve the global structure of the images, we opt to resize rather than crop them. Moreover, to reduce computational load, we down-sample the 2D images to a common smallest size (256×256) using bicubic interpolation, and then use these resized images for CycleGAN training.

Based on previous work [7], we choose a residual neural network (ResNet) as the generator instead of a U-net, while employing the default PatchGAN as the discriminator. The weights for the adversarial, cycle consistency, and identity losses are set to 1:10:5, respectively. Training is conducted with the Adam optimizer over 200 epochs: the first 100 epochs use a fixed learning rate of 0.00015, and the following 100 epochs apply a linear decay. A batch size of 10 is used throughout. Once training is complete, the translated ceT1 scans become hrT2 scans, allowing the segmentation model to be trained in a supervised manner using these generated images.

### 2.2 Self-training

We train the segmentation model in a supervised manner using the generated pseudo hrT2 scans and their corresponding annotations to segment VS and cochlea in real hrT2 scans. Following previous work [8], we adopt the default 3D full-resolution U-Net configuration from the nnU-Net [9] framework for both training and inference. nnU-Net streamlines the critical decisions needed to build a successful segmentation pipeline by automatically configuring preprocessing, network architecture, and training parameters for any given dataset [9, 6]. Therefore, we maintain all of nnU-Net's automated settings without modification. To reduce training time (given that nnU-Net's default 5-fold cross-validation combined with self-training can be time-consuming), we trained a single segmentation model using all available data.

To further address potential performance drops caused by subtle distribution differences between generated and real hrT2 scans, we employ a self-training strategy. This approach uses segmentation predictions from real hrT2 scans as pseudo labels, thereby creating a combined dataset that includes pseudo hrT2 scans paired with real ceT1 annotations and real hrT2 scans paired with predicted pseudo labels. The segmentation network is then retrained on this combined dataset, and new pseudo labels are generated for the real hrT2 scans to construct an updated dataset. Previous studies [5, 7] have demonstrated that this iterative self-training strategy improves model per-formance. Based on the observed performance on the crossMoDA 2022 challenge validation set, we set the number of self-training iterations to 3.

## 3  Results

All models were implemented in PyTorch 1.10 and both training and inference were performed on an RTX 3090 GPU with 24GB of memory. The training data were entirely sourced from the crossMoDA 2022 challenge set, and all results were obtained through its verification phase leaderboard. Table 1 presents the segmentation outcomes of various methods.

**Table 1.** Comparison of results on the validation set between different segmentation models.

| Methods | DSC (%) ↑ | | | ASSD (mm) ↓ | |
| --- | --- | --- | --- | --- | --- |
| | Mean | VS | Cochlea | VS | Cochlea |
| U-net w/o ST | 0.6734±0.1613 | 0.5455±0.2969 | 0.8012±0.0663 | 10.2882±9.4307 | 0.7161±2.6539 |
| ResNet w/o ST | 0.7908±0.0908 | 0.7890±0.1688 | 0.7927±0.0319 | 3.2845±5.8117 | 0.2594±0.1690 |
| ResNet w/ ST iter1 | 0.8086±0.0808 | 0.8114±0.1639 | 0.8058±0.0269 | 2.6269±3.6986 | 0.2399±0.1623 |
| ResNet w/ ST iter2 | 0.8211±0.0592 | 0.8337±0.1089 | 0.8085±0.0249 | 1.7864±2.1005 | 0.2346±0.1603 |
| ResNet w/ ST iter3 | **0.8225±0.0613** | **0.8351±0.1152** | **0.8098±0.0233** | **1.6712±2.1948** | **0.2317±0.1577** |

**Table 1** demonstrates that the ResNet-based generator outperforms the U-Net-based generator in CycleGAN image translation. Moreover, incorporating a self-training strategy enables the segmentation network to better adapt to real hrT2 scans, leading to enhanced segmentation performance. Specifically, on the verification phase leaderboard of the crossMoDA 2022 challenge, our final method achieved a DSC of 0.8351 ± 0.1152 and an ASSD of 1.6712 ± 2.1948 for vestibular schwannoma (VS), and a DSC of 0.8098 ± 0.0233 and an ASSD of 0.2317 ± 0.1577 for the cochlea. With an average DSC of 0.8225 across both structures, these results demonstrate that publicly available deep learning methods (CycleGAN and nnU-Net) can deliver robust performance on unsupervised cross-modality domain adaptation tasks without the need for structural modifications or specialized network designs.